\documentclass[letterpaper, 10 pt, conference]{ieeeconf}  % Comment this line out if you need a4paper

\IEEEoverridecommandlockouts                              % This command is only needed if 
\usepackage{amsmath,amsfonts}
\usepackage{algorithmic}
\usepackage{algorithm}
\usepackage{array}
\usepackage[caption=false,font=footnotesize]{subfig}
\usepackage{textcomp}
\usepackage{stfloats}
\usepackage{url}
\usepackage{verbatim}
\usepackage{graphicx}
\usepackage{cite}
\usepackage[acronym]{glossaries}
\newacronym{uav}{UAV}{Unmanned Aerial Vehicle}

\newacronym{ekf}{EKF}{Extended Kalman Filter}
\newacronym{esc}{ESC}{Electronic Speed Controller}
\newacronym{ukf}{UKF}{Unscented Kalman Filter}
\newacronym{imu}{IMU}{Inertial Measurement System}
\newacronym{gps}{GPS}{Global Positioning System}
\newacronym{indi}{INDI}{Incremental Nonlinear Dynamic Inversion}
\newacronym{tcn}{TCN}{Temporal Convolutional Network}
\newacronym{rpm}{RPM}{Revolutions per Minute}
\newacronym{bem}{BEM}{Blade Element Momentum}
\newacronym{rmse}{RMSE}{Root Mean Square Error}
\newacronym{lasso}{LASSO}{Least Absolute Shrinkage and Selection Operator}
\newacronym{rps}{RPS}{revolutions per second}
\newacronym{ned}{NED}{North-East-Down}
\newacronym{ls}{LS}{Least Squares}
\newacronym{nrmse}{nRMSE}{normalized Root Mean Square Error}
\newacronym{rls}{RLS}{Recursive Least Squares}
\newacronym{tow}{TOW}{Take-off Weight}
\newacronym{gvf}{GVF}{Guiding Vector Field}
\newacronym{vtol}{VTOL}{Vertical Takeoff and Landing}
\newacronym{pwm}{PWM}{Pulse Width Modulation}

\usepackage{booktabs}   % toprule midrule etc for tables
\usepackage{makecell}
\usepackage{multirow}

\usepackage{overpic}
\usepackage{tikz}
\usepackage{tikz-3dplot}
\usetikzlibrary{arrows.meta}

\usepackage{float}
\usepackage{xcolor}

\usepackage{siunitx}
\DeclareSIUnit{\rpm}{RPM}

\makeatletter
\let\NAT@parse\undefined
\makeatother
\usepackage[hidelinks]{hyperref}
\glsdisablehyper

\usepackage[nameinlink,capitalise]{cleveref}  
\crefname{equation}{}{} % remove "eq." before number
\title{\LARGE \bf
Coordinated Incremental Trajectory Tracking of a Tailsitter Drone
}

\author{Evangelos Ntouros$^{*}$ and Ewoud J. J. Smeur% <-this % stops a space
\thanks{This work has been submitted to the IEEE for possible publication. Copyright may be transferred without notice, after which this version may no longer be accessible.}% <-this % stops a space
\thanks{All authors are with Faculty of Aerospace Engineering, Delft University
of Technology, The Netherlands.}%
\thanks{$^{*}$Corresponding author
        {\tt\small e.ntouros@tudelft.nl}}%
}

\begin{document}

\maketitle
\thispagestyle{empty}
\pagestyle{empty}

%%%%%%%%%%%%%%%%%%%%%%%%%%%%%%%%%%%%%%%%%%%%%%%%%%%%%%%%%%%%%%%%%%%%%%%%%%%%%%%%
\begin{abstract}
This paper derives an analytical differential flatness transform for a tailsitter \gls{uav} under coordinated flight conditions using a simplified aerodynamic model. The proposed framework is formulated exclusively using rotation matrices, avoiding the ambiguities inherent to Euler angle representations. The method extends the applicability of an existing state-of-the-art differential flatness-based controller to flight regimes involving a significant vertical velocity component, where the previous approach becomes inapplicable. The proposed framework is validated experimentally with trajectories that highlight its advantages in these regimes.
\end{abstract}

\glsresetall

%%%%%%%%%%%%%%%%%%%%%%%%%%%%%%%%%%%%%%%%%%%%%%%%%%%%%%%%%%%%%%%%%%%%%%%%%%%%%%%%
\section{Introduction}
Hybrid \glspl{uav} combine the hover capability of rotorcraft with the higher cruise speed and aerodynamic efficiency of fixed-wing aircraft. A specific subclass is the tailsitter \gls{uav}, which transitions from hover to forward flight by pitching down approximately $90^\circ$. Typical tailsitter configurations include the dual-motor tailsitter, equipped with two non-tilting propellers and two elevons \cite{bronz2017}, and the quadrotor tailsitter, which integrates a quadrotor configuration with a lifting wing \cite{gu2018}. Due to their simplicity, tailsitters have attracted considerable attention in recent years for applications such as search and rescue, mapping, and package delivery. However, controlling these vehicles remains challenging as they combine coordinated and uncoordinated flight and can operate across a wide flight envelope including the post-stall regime during transition flight.

Tal et al. \cite{tal2022} employed a dynamic model of a dual-motor tailsitter based on the $\phi$-theory parameterization proposed in \cite{lustosa2019}. This formulation enabled them to prove the differential flatness property of the platform for uncoordinated flight. Leveraging this property, they computed feedforward angular rate signals and integrated them within the incremental control architecture of \cite{ewoud2020}, demonstrating agile uncoordinated flight. Although coordinated flight was also considered, the presented results relied on assumptions valid only for level flight and therefore do not directly extend to trajectories with a significant vertical velocity component.

To address coordinated flight with vertical velocity component, authors in \cite{lu2024} demonstrated the differential flatness property of a quadrotor tailsitter for coordinated flight conditions. Their approach relies on the standard Buckingham $\pi$-theory aerodynamic model identified from wind-tunnel experiments. While this formulation captures a broader range of aerodynamic effects, it requires a costly and time consuming identification process. Moreover, it introduces singularities around hover where the angle of attack and the sideslip angle are not defined. Furthermore, as shown in \cite{lu2024}, the differential flatness transform cannot be obtained in closed form and must instead be derived numerically, due to the increased complexity of the underlying aerodynamic model.

Researchers in \cite{wang2024} employed the simplified aerodynamic model of \cite{pucci2011} and derived an analytical differential flatness transform for the coordinated flight of quadcopter tailsitters with arbitrary wing installation angles. While extending the applicability to generic platform configurations, the formulation remains based on the angle of attack and the demonstrated results are limited to level flight.

The contribution of this work is the derivation of an analytical differential flatness transform for a tailsitter aircraft in coordinated flight, using the $\phi$-theory parameterization. The derivation is carried out for a quadrotor tailsitter configuration; however, the resulting framework can be extended to dual-motor tailsitters, as it relies on the same aerodynamic parameterization. The attitude and rotational dynamics are expressed exclusively with rotation matrices, to avoid the singularities and ambiguities of the Euler angle representations. The validity of the proposed approach is demonstrated experimentally\footnote{Autopilot code: https://github.com/tudelft/paparazzi/tree/swing} using trajectories with significant vertical velocity component. The experiments highlight the operating conditions under which the method of \cite{tal2022} becomes unsuitable and leads to singularities. By contrast, the proposed transform remains well-defined under these conditions. We view the proposed framework as complementary to \cite{tal2022}: the proposed approach addresses the coordinated flight regime, whereas formulation in \cite{tal2022} addresses uncoordinated flight.

\section{Tailsitter model}
\label{sec:model}
\subsection{Target vehicle and reference frames}

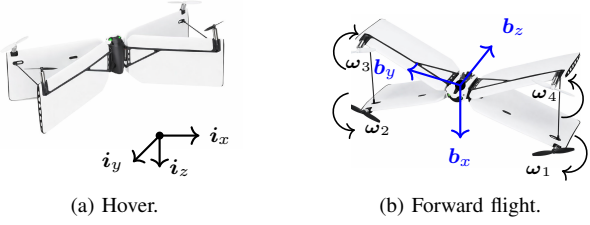
\begin{figure}[t]
    \centering

    \subfloat[Hover.\label{swing_hover}]{
      \begin{tikzpicture}[x=1pt,y=1pt]
        \node[inner sep=0, anchor=south west] (img)
          {\includegraphics[width=0.39\columnwidth]{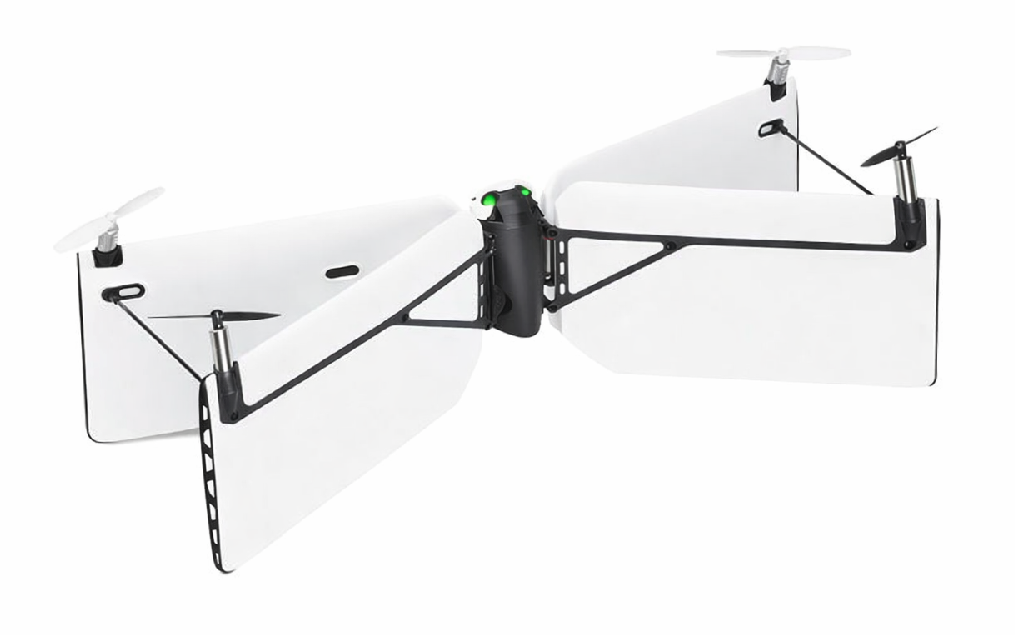}};

        \begin{scope}[shift={(img.south west)}]
          % Inertial frame origin
          \coordinate (I) at (65,5);
          \fill (I) circle (1.5pt);

          % Inertial axes
          \draw[->, line width=0.8pt] (I) -- ++(15,0)
            node[right] {\footnotesize $\boldsymbol{i}_x$};
          \draw[->, line width=0.8pt] (I) -- ++(0,-12)
            node[right] {\footnotesize $\boldsymbol{i}_z$};
          \draw[->, line width=0.8pt] (I) -- ++(-10,-10)
            node[left] {\footnotesize $\boldsymbol{i}_y$};
        \end{scope}
      \end{tikzpicture}
    }\hspace{0.8cm}
    \subfloat[Forward flight.\label{swing_fw}]{
      \begin{tikzpicture}[x=1pt,y=1pt]
        \node[inner sep=0, anchor=south west] (img)
          {\includegraphics[width=0.35\columnwidth]{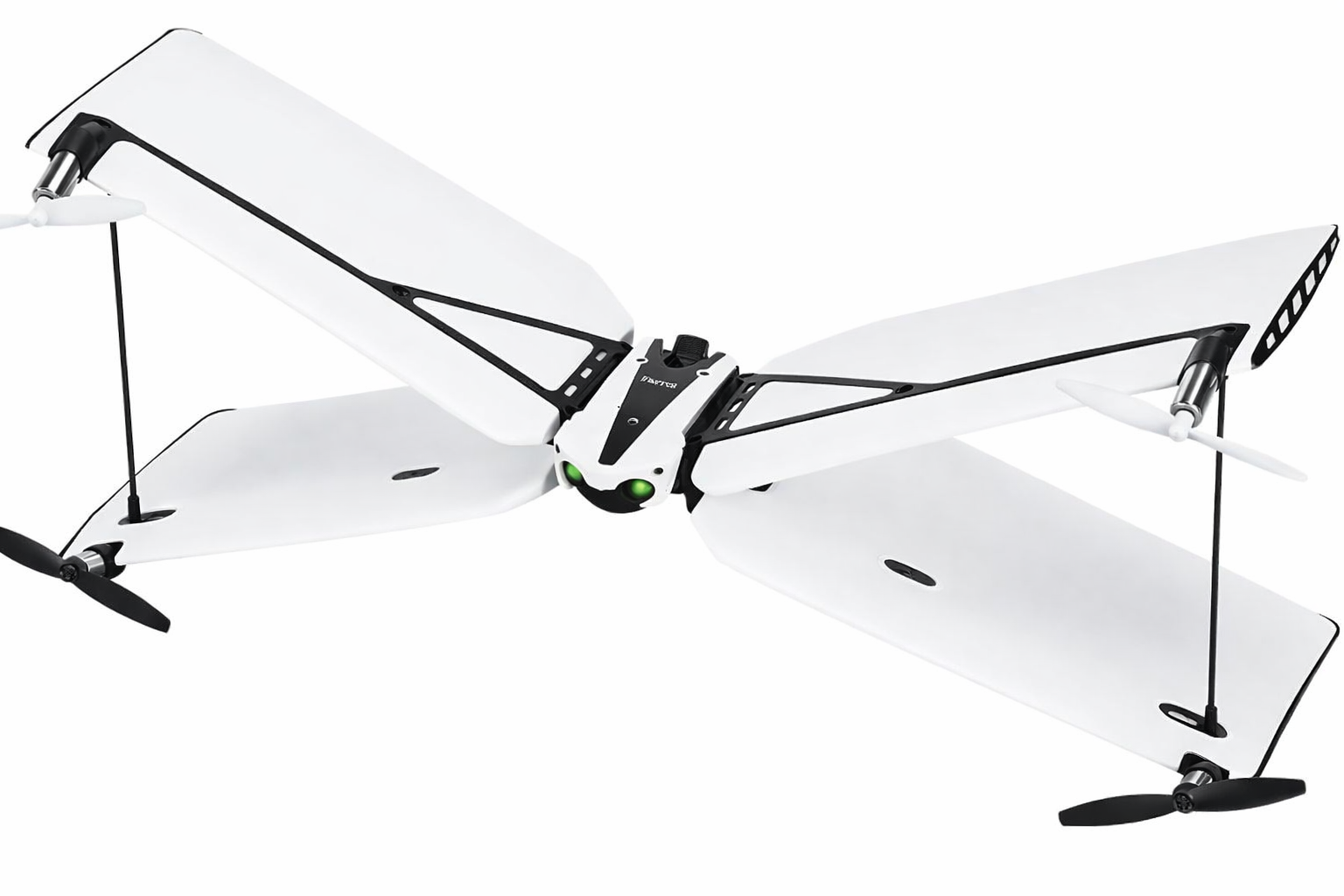}};

        \begin{scope}[shift={(img.south west)}]
          % Center of mass
          \coordinate (O) at (41,31);
          \fill (O) circle (2.0pt);

          % Body axes
          \draw[->, blue, line width=0.8pt] (O) -- ++(-20, 6)
            node[left] {\footnotesize $\boldsymbol{b}_y$};
          \draw[->, blue, line width=0.8pt] (O) -- ++(0, -20)
            node[below] {\footnotesize $\boldsymbol{b}_x$};
          \draw[->, blue, line width=0.8pt] (O) -- ++(12, 15)
            node[above right] {\footnotesize $\boldsymbol{b}_z$};

          % Motor locations
          \coordinate (M1) at (-8,37);
          \coordinate (M2) at (85,19);
          \coordinate (M3) at (-7,28);
          \coordinate (M4) at (90,12);

          % Curved arrows indicating propeller rotation
          \draw[->, semithick]
            ($(M1)+(6,0)$) arc[start angle=250,end angle=70,radius=7]
            node[below,yshift=-6pt] {\scriptsize $\boldsymbol{\omega}_3$};

          \draw[->, semithick]
            ($(M2)+(-6,0)$) arc[start angle=-100,end angle=80,radius=7]
            node[below left] {\scriptsize $\boldsymbol{\omega}_4$};

          \draw[->, semithick]
            ($(M3)+(6,0)$) arc[start angle=100,end angle=280,radius=7]
            node[right] {\scriptsize $\boldsymbol{\omega}_2$};

          \draw[->, semithick]
            ($(M4)+(-6,0)$) arc[start angle=70,end angle=-110,radius=7]
            node[left] {\scriptsize $\boldsymbol{\omega}_1$};
        \end{scope}
      \end{tikzpicture}
    }

    \caption{Parrot Swing tailsitter and reference frames.}
    \label{photo_swing}
\end{figure}

The target vehicle is the Parrot Swing, the quadrotor tailsitter shown in \cref{photo_swing}. It comprises an X-shaped lifting wing and four inward-tilted motors, a configuration that reduces collective thrust but increases yaw authority. 

We consider the Earth-fixed \gls{ned} inertial frame $\mathcal{I}$ with basis vectors $\{\boldsymbol{i}_x,\boldsymbol{i}_y,\boldsymbol{i}_z\}$ and the body-fixed frame $\mathcal{B}$ with basis vectors $\{\boldsymbol{b}_x,\boldsymbol{b}_y,\boldsymbol{b}_z\}$, expressed in the inertial frame. Following a hover-based convention, $\boldsymbol{b}_z$ is aligned with the vehicle longitudinal axis, $\boldsymbol{b}_y$ with the lateral axis, and $\boldsymbol{b}_x=\boldsymbol{b}_y\times\boldsymbol{b}_z$, such that $\boldsymbol{b}_x$ points forward during hover. The attitude of the vehicle is represented by the rotation matrix $R_b^i = 
    \begin{bmatrix}
        \boldsymbol{b}_x & \boldsymbol{b}_y & \boldsymbol{b}_z
    \end{bmatrix}
    \in SO(3)$, 
or equivalently, by the normalized quaternion $\boldsymbol{q} = \begin{bmatrix}
        q_s & \boldsymbol{q}_v
    \end{bmatrix}^\top
$, where $q_s$ is the scalar part and $\boldsymbol{q}_v$ the vector part.

\subsection{Equations of motion}
The translational dynamics of the vehicle are given by
\begin{subequations}\label{eq:transl_dyn}
\begin{align}
    \dot{\boldsymbol{p}} &= \boldsymbol{v} \label{eq:xdot} \\
    \dot{\boldsymbol{v}}
        &= {R_b^i}\boldsymbol{f}^b(\boldsymbol{v}_a,\boldsymbol{u}) + \boldsymbol{g},
        \label{eq:vdot}
\end{align}
\end{subequations}
where $\boldsymbol{p}$ and $\boldsymbol{v}$ denote, respectively, the position and velocity of the vehicle expressed in the inertial frame, and $\boldsymbol{v}_a = \boldsymbol{v} - \boldsymbol{v}_w$ is the velocity relative to the airmass, with $\boldsymbol{v}_w$ denoting the wind velocity. Wind is not considered in this work; nevertheless, the formulation is expressed in terms of $\boldsymbol{v}_a$ to retain a general form. The control input is $\boldsymbol{u}~=~    \begin{bmatrix}
        \omega_1 & \omega_2 & \omega_3 & \omega_4
    \end{bmatrix}^\top,$ where $\omega_i$ with $i \in \{1,2,3,4\}$ denotes the rotational speed of the i-th propeller as depicted in \cref{photo_swing}. The net specific force is denoted by $\boldsymbol{f}$\footnote{Vector superscripts denoting reference frames and function arguments may be omitted when clear from context.}. The rotational dynamics are given by
\begin{subequations}\label{eq:rot_dyn}
\begin{align}
    \dot{\boldsymbol{q}} &= \frac{1}{2}\boldsymbol{q}\otimes\boldsymbol{\Omega} \\
    \dot{\boldsymbol{\Omega}}
        &= J^{-1}\boldsymbol{M}(\boldsymbol{u}) - J^{-1}\boldsymbol{\Omega} \times J \boldsymbol{\Omega},
        \label{eq:Omegadot}
\end{align}
\end{subequations}
where $\boldsymbol{\Omega}$ is the angular rate of the vehicle, $\boldsymbol{M}$ is the net moment, and $J$ is the moment of inertia matrix. The symbol $\otimes$ denotes the Hamilton quaternion product \cite{stevens-lewis}. 

\subsection{$\phi$-theory parameterization and identification}
We employ the $\phi$-theory model \cite{lustosa2019} to express the aerodynamic force and moment acting on the vehicle. This parameterization offers two key advantages over the $\pi$-theory formulation. First, it provides a global model applicable across the broad flight envelope of the tailsitter, including transition phases and post-stall behavior. The model can be identified directly from experimental flight data, avoiding the costly identification of lift and drag coefficients. Second, it avoids the singularities inherent to the $\pi$-theory formulation, present in near-hover conditions when the angle of attack $\alpha$ and sideslip $\beta$ are not defined. The net specific force in the body frame is modeled as
\begin{equation}
    \boldsymbol{f}^b(\boldsymbol{v}_a,\boldsymbol{u}) 
    = \boldsymbol{f}^b_a(\boldsymbol{v}_a) + \boldsymbol{\tau}^b(\boldsymbol{u}),
    \label{eq:fbxyz}
\end{equation}
where the net aerodynamic force $\boldsymbol{f}_a$ and the specific thrust $\boldsymbol{\tau}$ are modeled respectively as
% \begin{equation}
%     \boldsymbol{f}_a^b =
%     \begin{bmatrix}
%         c_x \|\boldsymbol{v}_a\| \boldsymbol{i}_x^\top \boldsymbol{v}_a^b &
%         0 &
%         c_z \|\boldsymbol{v}_a\| \boldsymbol{i}_z^\top \boldsymbol{v}_a^b
%     \end{bmatrix}^\top,
% \end{equation}
% \begin{equation}
%     \boldsymbol{\tau}(\boldsymbol{u}) =
%     \begin{bmatrix}
%         0 &
%         0 &
%         c_\tau \sum_{i=1}^{4} \omega_i^2
%     \end{bmatrix}^\top.
%     \label{eq:spec_thrust}
% \end{equation}
\begin{subequations}
    \begin{align}
    \boldsymbol{f}_a^b(\boldsymbol{v}_a) &=
    \begin{bmatrix}
        c_x \|\boldsymbol{v}_a\| \boldsymbol{i}_x^\top \boldsymbol{v}_a^b &
        0 &
        c_z \|\boldsymbol{v}_a\| \boldsymbol{i}_z^\top \boldsymbol{v}_a^b
    \end{bmatrix}^\top \\
    \boldsymbol{\tau}^b(\boldsymbol{u}) &=
    \begin{bmatrix}
        0 &
        0 &
        \tilde{c}_{\tau} \sum_{i=1}^{4} \omega_i^2
    \end{bmatrix}^\top.
    \label{eq:spec_thrust}
    \end{align}
\end{subequations}

For the rotational dynamics we assume a diagonal moment of inertia matrix $J \in \mathbb{R}^{3 \times 3}
$, then if we set $\boldsymbol{m}(\boldsymbol{u})~=~J^{-1}\boldsymbol{M}(\boldsymbol{u})$ and $\boldsymbol{m}_\textrm{cor} = J^{-1}\boldsymbol{\Omega} \times J \boldsymbol{\Omega}$, we can write the equation \cref{eq:Omegadot} in the following compact form:
\begin{equation}
    \dot{\boldsymbol{\Omega}} = \boldsymbol{m}(\boldsymbol{u}) - \boldsymbol{m}_\textrm{cor},
    \label{eq:mb}
\end{equation}
where
\begin{equation}
    \boldsymbol{m}(\boldsymbol{u}) = \begin{bmatrix}
    \mu_x \left( \omega_1^2 -\omega_2^2 -\omega_3^2 + \omega_4^2 \right) \\
    \mu_y \left( \omega_1^2 +\omega_2^2 -\omega_3^2 - \omega_4^2 \right) \\
    \mu_z \left( -\omega_1^2 +\omega_2^2 -\omega_3^2 + \omega_4^2 \right) \\
    \end{bmatrix}
    \label{eq:mbxyz}.
\end{equation}
This formulation enables identification of the model parameters without explicit knowledge of the the moment of inertia matrix, as this is embedded within the $\mu_j$ with $j\in\{x,y,z\}$ coefficients.

We note that the formulation in \cref{eq:fbxyz,eq:mbxyz} corresponds to a simplified representation of the full $\phi$-theory model. The identification dataset was obtained from indoor flights, which limited the range of feasible maneuvers and the excitation of some state-dependent dynamics, such as moments induced by airspeed and rotational rates. To improve model generalizability, small or inconsistent terms were therefore neglected. Nevertheless, the unmodelled state-dependent effects are comparatively slow changing relative to the actuator-related terms and are thus appropriately handled by the incremental controller presented later, as demonstrated in~\cite{tal2022}. 

\Cref{tab:swing_coeff} lists the identified coefficients and the physical meaning of the associated terms. Since no direct measurements of the propeller rotational speeds are available, they are estimated using the first order actuator model \cite{ewoud2020}
$
    \dot{\boldsymbol{u}}
    ~=~
    -\omega_c
    \left(
    \boldsymbol{u}
    ~-~
    \boldsymbol{u}_c
    \right),
$
where $\boldsymbol{u}_c\in[0,1]$ is the normalized control signal sent to the actuators. The cutoff frequency was identified through bench tests as $\omega_c = 15~\mathrm{rad/s}$.
\begin{table}[h]
    \centering
    \caption{Parrot Swing model coefficients}
    % \scriptsize
    \begin{tabular}{c p{0.26\columnwidth} S[table-format=1.2e1] l}
        \toprule
        {Parameter} & {Physical meaning} & {Value} & {Units}\\
        \midrule
        $c_x$ & lift& -1.11e0 & $\mathrm{1/m}$ \\ 
        $c_z$ & drag& -1.54e-1 & $\mathrm{1/m}$ \\ 
        $\tilde{c}_{\tau}$ & effective thrust (due to tilted motors) & -4.42e-1 & $\mathrm{m/s^{2} }$\\ 
        $\mu_x$ & moment about $\boldsymbol{b}_x$  & 3.56e0 & $\mathrm{rad/s^{2} }$ \\
        $\mu_y$ & moment about $\boldsymbol{b}_y$ & 8.05e0 & $\mathrm{rad/s^{2} }$\\
        $\mu_z$ & moment about $\boldsymbol{b}_z$  & 7.84e-1 & $\mathrm{rad/s^{2} }$\\
        \midrule
    \end{tabular}
    \label{tab:swing_coeff}
\end{table}

\section{Differential Flatness Transform for coordinated flight}
\label{sec:diff_flatness}
In this section, we derive the differential flatness transform for the tailsitter model presented in \cref{sec:model}, given the coordinated flight constraint. The flat output is the vehicle position $\boldsymbol{p}$, assumed to be four times continuously differentiable.

\subsection{Attitude and specific thrust}
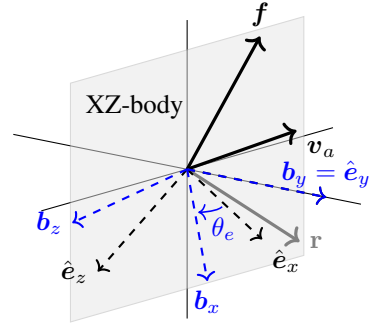
\begin{figure}[h]
    \centering
    \tdplotsetmaincoords{70}{40}
    \begin{tikzpicture}[tdplot_main_coords, scale=3, yscale=0.7,
        axis/.style={-, thin},
        vec/.style={->, very thick},
        proj/.style={->, thick, dashed},
        helper/.style={dashed, gray},
        bvec/.style={->, thick, dashed, blue}
    ]

        % Parameters
        \def\L{0.8}       % length of e- and b-frame axes
        \def\Lr{0.7}      % length of r
        \def\gwnia{50}
        \def\thetae{30}
        \def\thetaaa{15}

        % Origin
        \coordinate (O) at (0,0,0);

        % e-frame axes
        \pgfmathsetmacro{\exY}{\L*cos(\gwnia)}
        \pgfmathsetmacro{\exZ}{-\L*sin(\gwnia)}
        \pgfmathsetmacro{\ezY}{\L*sin(\gwnia)}
        \pgfmathsetmacro{\ezZ}{\L*cos(\gwnia)}

        \coordinate (Ex) at (0,\exY,\exZ);
        \coordinate (Ey) at (\L,0,0);
        \coordinate (Ez) at (0,-\ezY,-\ezZ);

        % b-frame axes: rotated by theta_e with respect to e-frame
        \pgfmathsetmacro{\btheta}{\gwnia+\thetae}

        \pgfmathsetmacro{\bxY}{\L*cos(\btheta)}
        \pgfmathsetmacro{\bxZ}{-\L*sin(\btheta)}
        \pgfmathsetmacro{\bzY}{\L*sin(\btheta)}
        \pgfmathsetmacro{\bzZ}{\L*cos(\btheta)}

        \coordinate (Bx) at (0,\bxY,\bxZ);
        \coordinate (By) at (\L,0,0);
        \coordinate (Bz) at (0,-\bzY,-\bzZ);

        % Vector r
        \coordinate (R) at (0,\Lr*1.1,-\Lr/1);

        % Axes
        \draw[axis] (-1,0,0) -- (1,0,0);
        \draw[axis] (0,-1,0) -- (0,1,0);
        \draw[axis] (0,0,-1) -- (0,0,1);

% XZ-body plane
        \filldraw[fill=gray!20, draw=gray!60, opacity=0.5]
            (0,-0.8,-0.8) -- (0,0.8,-0.8) -- (0,0.8,0.8) -- (0,-0.8,0.8) -- cycle;
        
        \node at (0,-0.36,0.52) {$\text{XZ-body}$};

        % Keep these as in your original figure
        \draw[vec, black] (O) -- (0,0.75,0.05) node[below right] {$\boldsymbol{v}_a$};
        \draw[vec, black] (O) -- (0,0.5,0.75) node[above] {$\boldsymbol{f}$};

        % Vector r
        \draw[vec, gray] (O) -- (R) node[right] {$\mathbf r$};

        % e-frame axes
        \draw[proj, black] (O) -- (Ex) node[below right] {$\hat{\boldsymbol e}_x$};
        \draw[proj, black] (O) -- (Ey) node[below] {};
        \draw[proj, black] (O) -- (Ez) node[left] {$\hat{\boldsymbol{e}}_z$};

        % b-frame axes
        \draw[bvec] (O) -- (Bx) node[below] {$\boldsymbol {b}_x$};
        \draw[bvec] (O) -- (By) node[above] {$\boldsymbol {b}_y=\hat{\boldsymbol {e}}_y$};
        \draw[bvec] (O) -- (Bz) node[left] {$\boldsymbol {b}_z$};

        \def\rang{0.5}
        \draw[->,blue,semithick]
            ((0,{\rang*sin(\thetaaa)+0.11},{\rang*cos(\thetaaa)-0.8})
            arc[
                start angle={30},
                end angle={-40},
                radius=0.13
            ];
        \node[blue] at (0,0.25,-0.47){$\theta_e$};

        % Helper projection lines for r
        % \draw[helper] (R) -- (\Lr,\Lr,0);
        % \draw[helper] (\Lr,0,0) -- (\Lr,\Lr,0);
        % \draw[helper] (0,\Lr,0) -- (\Lr,\Lr,0);
        % \draw[helper] (R) -- (0,\Lr,\Lr);
        % \draw[helper] (0,\Lr,\Lr) -- (0,0,\Lr);
        % \draw[helper] (0,\Lr,\Lr) -- (0,\Lr,0);

    \end{tikzpicture}
    \caption{Attitude representation for coordinated flight flatness transform.}
    \label{fig:xz_plane}
\end{figure}

For coordinated flight of symmetric aircraft, the velocity relative to the airmass $\boldsymbol{v}_a$ and the specific force $\boldsymbol{f}$ must lie on the XZ plane of $\mathcal{B}$. Trivially, we obtain $\boldsymbol{v}_a=\dot{\boldsymbol{p}} -\boldsymbol{v}_w$ and  $\boldsymbol{f} = \ddot{\boldsymbol{p}} -\boldsymbol{g} $. Thus we derive the body-y axis by
\begin{equation}
    \boldsymbol{b}_y = s\frac{\boldsymbol{v}_a \times \boldsymbol{f}}{\| \boldsymbol{v}_a \times \boldsymbol{f} \|},
    \label{eq:by}
\end{equation}
where $s = \mathrm{sign}\left( 
\left( \boldsymbol{v}_a \times \boldsymbol{f} \right)
\cdot
\left( \boldsymbol{v}_{a,\mathrm{prev}} \times \boldsymbol{f}_{\mathrm{prev}} \right)
\right)$ with the subscript ``prev'' denoting the previous time step. This ensures attitude continuity; i.e., when flying backward, the vehicle transitions to inverted flight rather than yawing by $180^\circ$. Next we define an intermediate reference frame $\mathcal{E}$ with basis vectors 
$\{\hat{\boldsymbol{e}}_x, \hat{\boldsymbol{e}}_y, \hat{\boldsymbol{e}}_z\}$, such that $\hat{\boldsymbol{e}}_y=\boldsymbol{e}_y/\| \boldsymbol{e}_y\|=\boldsymbol{b}_y$ with $\boldsymbol{e}_y = s\left( \boldsymbol{v}_a \times \boldsymbol{f} \right)$. Let $\boldsymbol{r} \in \mathbb{R}^3$ be any constant vector not parallel to $\boldsymbol{e}_y$, then compute $\boldsymbol{e}_z$ as
\begin{equation}
    \boldsymbol{e}_z = \boldsymbol{r} \times \boldsymbol{e}_y,
\end{equation}
which lies on the XZ plane, as illustrated in \cref{fig:xz_plane}. Then, the right-hand coordinate system is completed with
\begin{equation}
    \boldsymbol{e}_x = \boldsymbol{e}_y \times \boldsymbol{e}_z.
\end{equation}
We normalize to get $\hat{\boldsymbol{e}}_x = \boldsymbol{e}_x/\|\boldsymbol{e}_x\|$ and $\hat{\boldsymbol{e}}_z = \boldsymbol{e}_z/\|\boldsymbol{e}_z\|$, and thereby construct the rotation matrix that describes the rotation from the inertial frame $\mathcal{I}$ to the intermediate $\mathcal{E}$:
\begin{equation}
    R_i^e = \begin{bmatrix}
        \hat{\boldsymbol{e}}_x^\top & \hat{\boldsymbol{e}}_y^\top & \hat{\boldsymbol{e}}_z^\top
    \end{bmatrix}^\top.
    \label{eq:Re_i}
\end{equation}
Next, we construct the rotation matrix $R_e^b$ that represents the
rotation from frame $\mathcal{E}$ to the frame $\mathcal{B}$.
This corresponds to a rotation about $\boldsymbol{b}_y$
by an angle $\theta_e$ and is given by
\begin{equation}
    R_e^b = \begin{bmatrix}
        \cos\theta_e & 0 & -\sin\theta_e \\
        0 & 1 &0 \\
        \sin\theta_e &0 & \cos\theta_e
    \end{bmatrix}.
    \label{eq:Rb_e}
\end{equation}
We express the net specific force in the body frame as $\boldsymbol{f}^b~=~R_e^b \boldsymbol{f}^e$ and using \cref{eq:fbxyz} we expand to
\begin{equation}
\begin{aligned}
    c_x \| \boldsymbol{v}_a\| \cos\theta_e \boldsymbol{i}_x^\top \boldsymbol{v}_a^e
    &- c_x \| \boldsymbol{v}_a\| \sin\theta_e \boldsymbol{i}_z^\top \boldsymbol{v}_a^e
    = \\
    &\cos\theta_e \boldsymbol{i}_x^\top \boldsymbol{f}^e
    - \sin\theta_e \boldsymbol{i}_z^\top \boldsymbol{f}^e
    \label{eq:E:attitude}
\end{aligned}
\end{equation}
\begin{equation}
\begin{aligned}
    c_z \| \boldsymbol{v}_a\| \sin\theta_e \boldsymbol{i}_x^\top \boldsymbol{v}_a^e 
    &+ c_z \| \boldsymbol{v}_a\| \cos\theta_e \boldsymbol{i}_z^\top \boldsymbol{v}_a^e + \boldsymbol{i}_z^\top \boldsymbol{\tau}
    = \\ 
    &\sin\theta_e \boldsymbol{i}_x^\top \boldsymbol{f}^e + \cos\theta_e \boldsymbol{i}_z^\top \boldsymbol{f}^e.
    \label{eq:E:thrust}
\end{aligned}
\end{equation}
By \cref{eq:E:attitude} and \cref{eq:E:thrust} we obtain the angle $\theta_e$ and the specific thrust respectively: 
\begin{equation}
    \tan \theta_e = \frac{\sigma_x}{\sigma_z} \Rightarrow \theta_e = \mathrm{atan2}(\sigma_x,\sigma_z) + \kappa \pi,
    \label{eq:thetae}
\end{equation}
where
\begin{subequations}\label{eq:sigmas}
\begin{align}
    \sigma_x &= - \boldsymbol{i}_x^\top \boldsymbol{f}^e + c_x \| \boldsymbol{v}_a\| \boldsymbol{i}_x^\top \boldsymbol{v}_a^e \\
    \sigma_z &= - \boldsymbol{i}_z^\top \boldsymbol{f}^e + c_x \| \boldsymbol{v}_a\| \boldsymbol{i}_z^\top \boldsymbol{v}_a^e,
\end{align}
\end{subequations}
and
\begin{equation}
\begin{aligned}
    \boldsymbol{i}_z^\top \boldsymbol{\tau} &= \sin\theta_e \boldsymbol{i}_x^\top \boldsymbol{f}^e + \cos\theta_e \boldsymbol{i}_z^\top \boldsymbol{f}^e \\ &- c_z \|\boldsymbol{v}_a\|(\sin\theta_e \boldsymbol{i}_x^\top \boldsymbol{v}_a^e +\cos\theta_e \boldsymbol{i}_z^\top \boldsymbol{v}_a^e).
\end{aligned}
\end{equation}
We select $\kappa \in \{0,1\}$ such that $\boldsymbol{i}_z^\top \boldsymbol{\tau} < 0$. Finally, the attitude can be reconstructed using \cref{eq:Re_i} and \cref{eq:Rb_e} in
\begin{equation}
    R_i^b = R_e^b R_i^e.
    \label{eq:flatness:R}
\end{equation}

\subsection{Angular rate}
We compute the scew-symmetric matrix $[\boldsymbol{\Omega}]_\times$ from the rotational kinematics and then the angular rate $\boldsymbol{\Omega}$ with 
\begin{align}
    \dot{R}_i^b = -[\boldsymbol{\Omega}]_\times R_i^b \Rightarrow
    [\boldsymbol{\Omega}]_\times = - \dot{R}_i^b \left( {R_i^b} \right)^\top
    \label{eq:Omegax}
\end{align}
\begin{equation}
    \boldsymbol{\Omega} =
    \begin{bmatrix}
        [\boldsymbol{\Omega}]_{\times,32} &
        [\boldsymbol{\Omega}]_{\times,13} &
        [\boldsymbol{\Omega}]_{\times,21}
    \end{bmatrix}^\top,
    \label{eq:flatness:Omega}
\end{equation}
where $[\boldsymbol{\Omega}]_{\times,ij}$ denotes the $(i,j)$-th entry of $[\boldsymbol{\Omega}]_\times$. By differentiation of \cref{eq:flatness:R} we get
\begin{equation}
    \dot{R}_i^b = \dot{R}_e^b R_i^e + R_e^b \dot{R}_i^e.
\end{equation}
Differentiation of \cref{eq:Re_i} yields
\begin{equation}
    \dot{R_i^e} = \begin{bmatrix}
        \dot{\hat{\boldsymbol{e}}}_x^\top & \dot{\hat{\boldsymbol{e}}}_y^\top & \dot{\hat{\boldsymbol{e}}}_z^\top
    \end{bmatrix}^\top,
    \label{eq:R_i_e_dot}
\end{equation}
where $\dot{\hat{\boldsymbol{e}}}_j$ with $j\in\{x,y,z\}$, is computed by
\begin{equation}
    \dot{\hat{\boldsymbol{e}}}_j = \frac{1}{\|\boldsymbol{e}_j\|} (\mathbb{I}_3 - \frac{\boldsymbol{e}_j\boldsymbol{e}_j^\top}{\|\boldsymbol{e}_j\|^2}) \dot{\boldsymbol{e}}_j,   
\end{equation}
% and
\begin{subequations}
    \begin{align}
        \dot{\boldsymbol{e}}_y &= s \left( \dot{\boldsymbol{v}}_a \times \boldsymbol{f} + \boldsymbol{v}_a \times \dot{\boldsymbol{f}} \right) \\
        \dot{\boldsymbol{e}}_z &= \boldsymbol{r} \times \dot{\boldsymbol{e}}_y \\
        \dot{\boldsymbol{e}}_x &= \dot{\boldsymbol{e}}_y \times \boldsymbol{e}_z + \boldsymbol{e}_y \times \dot{\boldsymbol{e}}_z.
    \end{align}
\end{subequations}
Trivially, we get $\dot{\boldsymbol{v}}_a = \ddot{\boldsymbol{p}}$ and $\dot{\boldsymbol{f}} = \boldsymbol{p}^{(3)}$. Subsequently, differentiation of \cref{eq:Rb_e} yields
\begin{equation}
    \dot{R}_e^b = \begin{bmatrix}
        -\sin\theta_e & 0 & -\cos\theta_e \\
        0 & 0 &0 \\
        \cos\theta_e &0 & -\sin\theta_e
    \end{bmatrix}\dot{\theta}_e.
    \label{eq:Re_b_dot}
\end{equation}
Then, $\dot{\theta}_e$ is computed by differentiation of \cref{eq:thetae} as
\begin{equation}
    \dot{\theta}_e = \frac{\dot{\sigma}_x \sigma_z - \sigma_x \dot{\sigma}_z}{\sigma_x^2 + \sigma_z^2},
    \label{eq:theta_dot}
\end{equation}
where by differentiation of \cref{eq:sigmas} we obtain
\begin{subequations}
    \label{eq:sigmas_dot}
    \begin{align}
        \dot{\sigma}_x &= -\boldsymbol{i}_x^\top \dot{\boldsymbol{f}}^e + c_x \left ( \dot{\|\boldsymbol{v}_a\|} \boldsymbol{i}_x^\top \boldsymbol{v}_a^e +\| \boldsymbol{v}_a\|\boldsymbol{i}_x^\top \dot{\boldsymbol{v}}_a^e  \right)\\
        \dot{\sigma}_z &= -\boldsymbol{i}_z^\top \dot{\boldsymbol{f}}^e + c_x \left ( \dot{\|\boldsymbol{v}_a\|} \boldsymbol{i}_z^\top \boldsymbol{v}_a^e +\| \boldsymbol{v}_a\|\boldsymbol{i}_z^\top \dot{\boldsymbol{v}}_a^e  \right),
    \end{align}
\end{subequations}
with
\begin{equation}
    \dot{\|\boldsymbol{v}_a\|} = \frac{\boldsymbol{v}_a^\top \dot{\boldsymbol{v}}_a}{\|\boldsymbol{v}_a\|},
\end{equation}
\begin{subequations}
    \begin{align}
         \dot{\boldsymbol{f}}^e &= \dot{R}_i^e\boldsymbol{f} + R_i^e \dot{\boldsymbol{f}} \\
         \dot{\boldsymbol{v}}_a^e &= \dot{R}_i^e\boldsymbol{v}_a + R_i^e \dot{\boldsymbol{v}}_a.
    \end{align}
\end{subequations}

\subsection{Angular acceleration}
\label{subsec:ang_accel}
% The derivation of the angular acceleration is straightforward but lengthy. Thus intermediate steps are omitted for brevity and due to space limitations. 
We differentiate \cref{eq:Omegax} to recover the angular acceleration $\dot{\boldsymbol{\Omega}}$ with:
\begin{equation}
    \dot{[\boldsymbol{\Omega}]}_{\times}
    =
    -\ddot{R}_i^b \left(R_i^b\right)^{\top}
    -
    \dot{R}_i^b \left(\dot{R}_i^b\right)^{\top}
\end{equation}
\begin{equation}
    \dot{\boldsymbol{\Omega}} =
    \begin{bmatrix}
        \dot{[\boldsymbol{\Omega}]}_{\times,32} &
        \dot{[\boldsymbol{\Omega}]}_{\times,13} &
        \dot{[\boldsymbol{\Omega}]}_{\times,21}
    \end{bmatrix}^\top,
    \label{eq:flatness:Omegadot}
\end{equation}
where
\begin{equation}
    \ddot{R}_i^b = \ddot{R}_e^b R_i^e + 2\dot{R}_e^b \dot{R}_i^e + R_e^b \ddot{R}_i^e.
\end{equation}
From \cref{eq:R_i_e_dot} we compute $\ddot{R}_i^e$ with
% \begin{equation}
% \left.
% \begin{aligned}
% \ddot{R}_i^e
% &=
% \begin{bmatrix}
% \ddot{\hat{\boldsymbol{e}}}_x^\top &
% \ddot{\hat{\boldsymbol{e}}}_y^\top &
% \ddot{\hat{\boldsymbol{e}}}_z^\top
% \end{bmatrix}^{\top}
% \\[0.5em]
% \ddot{\hat{\boldsymbol{e}}}_j
% &=
% \frac{1}{\|\boldsymbol{e}_j\|}\ddot{\boldsymbol{e}}_j
% -
% \frac{2}{\|\boldsymbol{e}_j\|^3}
% \left(\boldsymbol{e}_j^\top \dot{\boldsymbol{e}}_j\right)
% \dot{\boldsymbol{e}}_j
% \\
% &\quad
% -
% \frac{1}{\|\boldsymbol{e}_j\|^3}
% \left(
% \dot{\boldsymbol{e}}_j^\top \dot{\boldsymbol{e}}_j
% +
% \boldsymbol{e}_j^\top \ddot{\boldsymbol{e}}_j
% \right)
% \boldsymbol{e}_j \\
% &\quad+ 
% \frac{3}{\|\boldsymbol{e}_j\|^5}
% \left(
% \boldsymbol{e}_j^\top \dot{\boldsymbol{e}}_j
% \right)^2
% \boldsymbol{e}_j
% \\[0.5em]
% \ddot{\boldsymbol{e}}_y
% &=
% \ddot{\boldsymbol{v}}_a \times \boldsymbol{f}
% +
% 2\dot{\boldsymbol{v}}_a \times \dot{\boldsymbol{f}}
% +
% \boldsymbol{v}_a \times \ddot{\boldsymbol{f}}
% \\
% \ddot{\boldsymbol{e}}_x
% &=
% \boldsymbol{r} \times \ddot{\boldsymbol{e}}_y
% \\
% \ddot{\boldsymbol{e}}_z
% &=
% \ddot{\boldsymbol{e}}_y \times \boldsymbol{e}_x
% +
% 2\dot{\boldsymbol{e}}_y \times \dot{\boldsymbol{e}}_x
% +
% \boldsymbol{e}_y \times \ddot{\boldsymbol{e}}_x
% \end{aligned}
% \right\}
% \end{equation}
\begin{equation}
    \ddot{R}_i^e = \begin{bmatrix}
        \ddot{\hat{\boldsymbol{e}}}_x^\top & \ddot{\hat{\boldsymbol{e}}}_y^\top & \ddot{\hat{\boldsymbol{e}}}_z^\top
    \end{bmatrix}^\top,
\end{equation}
where $\ddot{\hat{\boldsymbol{e}}}_j$ with $j\in\{x,y,z\}$, is derived by
\begin{equation}
\begin{aligned}
    \ddot{\hat{\boldsymbol{e}}}_j &= \frac{1}{\| {\boldsymbol{e}_j} \| } \ddot{\boldsymbol{e}}_j - \frac{2}{\| \boldsymbol{e}_j \|^3} \left (  \boldsymbol{e}_j^\top \dot{\boldsymbol{e}}_j \right)\dot{\boldsymbol{e}}_j \\ &- \frac{1}{\| \boldsymbol{e}_j \|^3} \left( \dot{\boldsymbol{e}}_j^\top\dot{\boldsymbol{e}}_j
    + \boldsymbol{e}_j^\top \ddot{\boldsymbol{e}}_j \right)\boldsymbol{e}_j  + \frac{3}{\| \boldsymbol{e}_j \|^5}\left(  \boldsymbol{e}_j^\top \dot{\boldsymbol{e}}_j  \right )^2\boldsymbol{e}_j,
\end{aligned}
\end{equation}
\begin{subequations}
    \begin{align}
        \ddot{\boldsymbol{e}}_y &= s \left( \ddot{\boldsymbol{v}}_a \times \boldsymbol{f} + 2 \dot{\boldsymbol{v}}_a \times \dot{\boldsymbol{f}} + \boldsymbol{v}_a \times \ddot{\boldsymbol{f}} \right) \\
        \ddot{\boldsymbol{e}}_x &= \boldsymbol{r} \times \ddot{\boldsymbol{e}}_y \\
        \ddot{\boldsymbol{e}}_z &= \ddot{\boldsymbol{e}}_y \times \boldsymbol{e}_z + 2 \dot{\boldsymbol{e}}_y \times \dot{\boldsymbol{e}}_z + \boldsymbol{e}_y \times \ddot{\boldsymbol{e}}_z.
    \end{align}
\end{subequations}
Trivially, $\ddot{\boldsymbol{f}} = \boldsymbol{p}^{(4)}$ and $\ddot{\boldsymbol{v}}_a = \boldsymbol{p}^{(3)}$. By \cref{eq:Re_b_dot} $\ddot{R}_e^b$ is derived as
\begin{equation}
\begin{aligned}
    \ddot{R}_e^b &= \begin{bmatrix}
        -\cos \theta_e & 0 & \sin \theta_e\\
        0 & 0 & \\
        -\sin \theta_e & 0 & -\cos \theta_e
    \end{bmatrix} \dot{\theta}^2_e \\ &+ \begin{bmatrix}
        -\sin\theta_e & 0 & -\cos\theta_e \\
        0 & 0 &0 \\
        \cos\theta_e &0 & -\sin\theta_e
    \end{bmatrix}\ddot{\theta}_e.
    \end{aligned}
\end{equation}
Subsequently, $\ddot{\theta}_e$ is computed by differentiation of \cref{eq:theta_dot} as
\begin{subequations}
\begin{align}
\ddot{\theta}_e &= \frac{\dot{\nu}_x\nu_z - \nu_x\dot{\nu}_z}{ \nu_z^2},\\
    \nu_x &= \dot{\sigma}_x \sigma_z - \sigma_x \dot{\sigma}_z \\
    \nu_z &= \sigma_x^2 + \sigma_z^2 \\    \dot{\nu}_x &= \ddot{\sigma}_x \sigma_z - \sigma_x \ddot{\sigma}_z \\
    \dot{\nu}_z &= 2\sigma_x\dot{\sigma}_x  + 2\sigma_z \dot{\sigma}_z.
\end{align}
\end{subequations}
Differentiation of \cref{eq:sigmas_dot} yields:
\begin{subequations}
\begin{align}
\begin{split}
    \ddot{\sigma}_x
    &= -\boldsymbol{i}_x^\top \ddot{\boldsymbol{f}}^e
    + c_x \Big(
    \ddot{\|\boldsymbol{v}_a\|}
    \boldsymbol{i}_x^\top \boldsymbol{v}_a^e
    + 2\dot{\|\boldsymbol{v}_a\|}
    \boldsymbol{i}_x^\top \dot{\boldsymbol{v}}_a^e \\
    &\qquad\qquad\qquad\quad
    + \|\boldsymbol{v}_a\|
    \boldsymbol{i}_x^\top \ddot{\boldsymbol{v}}_a^e
    \Big)
    \end{split}
    \\
    \begin{split}
    \ddot{\sigma}_z
    &= -\boldsymbol{i}_z^\top \ddot{\boldsymbol{f}}^e
    + c_x \Big(
    \ddot{\|\boldsymbol{v}_a\|}
    \boldsymbol{i}_z^\top \boldsymbol{v}_a^e
    + 2\dot{\|\boldsymbol{v}_a\|}
    \boldsymbol{i}_z^\top \dot{\boldsymbol{v}}_a^e \\
    &\qquad\qquad\qquad\quad
    + \|\boldsymbol{v}_a\|
    \boldsymbol{i}_z^\top \ddot{\boldsymbol{v}}_a^e
    \Big),
\end{split}
\end{align}
\end{subequations}
where
% \begin{subequations}
% \begin{equation}
%     \ddot{\| \boldsymbol{v}_a \|} = \frac{\left ( \dot{\boldsymbol{v}}_a^\top\dot{\boldsymbol{v}}_a +  \boldsymbol{v}_a^\top \ddot{\boldsymbol{v}}_a \right) \| \boldsymbol{v}_a \| - \boldsymbol{v}_a^\top \dot{\boldsymbol{v}}_a \dot{\| \boldsymbol{v}_a \|} }{\| \boldsymbol{v}_a \|^2},
% \end{equation}
% \begin{align}
%     \ddot{\boldsymbol{f}}^e &= \ddot{R}_i^e \boldsymbol{f} + 2\dot{R}_i^a \boldsymbol{f} + R_i^e \ddot{\boldsymbol{f}}\\
%     \dot{\boldsymbol{v}}_a^e &= \ddot{R}_i^e \boldsymbol{v}_a + 2\dot{R}_i^a \boldsymbol{v}_a + R_i^e \ddot{\boldsymbol{v}}_a.
% \end{align}
% \end{subequations}
\begin{equation}
    \ddot{\| \boldsymbol{v}_a \|} = \frac{\left ( \dot{\boldsymbol{v}}_a^\top\dot{\boldsymbol{v}}_a +  \boldsymbol{v}_a^\top \ddot{\boldsymbol{v}}_a \right) \| \boldsymbol{v}_a \| - \boldsymbol{v}_a^\top \dot{\boldsymbol{v}}_a \dot{\| \boldsymbol{v}_a \|} }{\| \boldsymbol{v}_a \|^2},
\end{equation}
\begin{subequations}
\begin{align}
    \ddot{\boldsymbol{f}}^e &= \ddot{R}_i^e \boldsymbol{f} + 2\dot{R}_i^e \boldsymbol{f} + R_i^e \ddot{\boldsymbol{f}}\\
    \ddot{\boldsymbol{v}}_a^e &= \ddot{R}_i^e \boldsymbol{v}_a + 2\dot{R}_i^e \boldsymbol{v}_a + R_i^e \ddot{\boldsymbol{v}}_a.
\end{align}
\end{subequations}
\subsection{Control input}
We express \cref{eq:mbxyz,eq:spec_thrust} in the following compact notation and recover the control input:
\begin{equation}
    \begin{bmatrix}
        \boldsymbol{m} \\ \boldsymbol{i}_z^\top \boldsymbol{\tau}    
    \end{bmatrix} = G \boldsymbol{u}^{\circ 2} \Rightarrow \boldsymbol{u} = \left ( G^{-1} \begin{bmatrix}
        \boldsymbol{m} \\ \boldsymbol{i}_z^\top \boldsymbol{\tau}    
    \end{bmatrix} \right ) ^{\circ \frac{1}{2}},
    \label{eq:flatness:control_input}
\end{equation}
% where $(\cdot)^{\circ }$ denotes elementwise exponentiation and
where $(\cdot)^{\circ 2}$ and $(\cdot)^{\circ \frac12}$ denote the element-wise square and square root, respectively; $G$ is the control effectiveness matrix
\begin{equation}
    G = \begin{bmatrix}
        \mu_x & -\mu_x & -\mu_x & \mu_x\\
        \mu_y & \mu_y & -\mu_y & -\mu_y\\
        -\mu_z & \mu_x & -\mu_z & \mu_z\\
        \tilde{c}_{\tau} & \tilde{c}_{\tau} & \tilde{c}_{\tau} & \tilde{c}_{\tau}\\
    \end{bmatrix}.
\end{equation}

% \begin{subequations}
% \label{eq:flatness:control_input}
% \begin{align}
%     \omega_1 &= \frac{1}{2} \left ( \frac{\boldsymbol{i}_z^\top \boldsymbol{\tau}}{c_\tau} + 
%     \frac{\boldsymbol{i}_x^\top \boldsymbol{m}}{\mu_x} +
%     \frac{\boldsymbol{i}_y^\top \boldsymbol{m}}{\mu_y} -
%     \frac{\boldsymbol{i}_z^\top \boldsymbol{m}}{\mu_z} \right )^{\frac{1}{2}} \\
%     \omega_2 &= \frac{1}{2} \left ( \frac{\boldsymbol{i}_z^\top \boldsymbol{\tau}}{c_\tau} - 
%     \frac{\boldsymbol{i}_x^\top \boldsymbol{m}}{\mu_x} +
%     \frac{\boldsymbol{i}_y^\top \boldsymbol{m}}{\mu_y} +
%     \frac{\boldsymbol{i}_z^\top \boldsymbol{m}}{\mu_z} \right )^{\frac{1}{2}} \\
%     \omega_3 &= \frac{1}{2} \left ( \frac{\boldsymbol{i}_z^\top \boldsymbol{\tau}}{c_\tau} - 
%     \frac{\boldsymbol{i}_x^\top \boldsymbol{m}}{\mu_x} -
%     \frac{\boldsymbol{i}_y^\top \boldsymbol{m}}{\mu_y} -
%     \frac{\boldsymbol{i}_z^\top \boldsymbol{m}}{\mu_z} \right )^{\frac{1}{2}} \\
%     \omega_4 &= \frac{1}{2} \left ( \frac{\boldsymbol{i}_z^\top \boldsymbol{\tau}}{c_\tau} + 
%     \frac{\boldsymbol{i}_x^\top \boldsymbol{m}}{\mu_x} -
%     \frac{\boldsymbol{i}_y^\top \boldsymbol{m}}{\mu_y} +
%     \frac{\boldsymbol{i}_z^\top \boldsymbol{m}}{\mu_z} \right )^{\frac{1}{2}}
%     \end{align}
% \end{subequations}

\subsection{Singularities}
\label{subsec:sing}
The presented differential flatness formulation exhibits a singularity condition when
$\| \boldsymbol{v}_a \times \boldsymbol{f} \| = 0$,
in which case the $\boldsymbol{b}_y$ axis cannot be uniquely defined using \cref{eq:by}. This condition can be further divided into three subcategories: 
\begin{enumerate}
    \item $\| \boldsymbol{v}_a \| = 0$ corresponding to near-hover conditions,
    \item $\| \boldsymbol{f} \| = 0$ corresponding to free fall, and
    \item $\boldsymbol{v}_a \parallel \boldsymbol{f}$ which may occur during takeoff and landing, as well as nose-dive and tail-slide maneuvers.
\end{enumerate}
We argue that singularity handling should primarily be addressed at the trajectory generation level, ensuring that only feasible trajectories are provided to the controller. Nevertheless, the controller should be capable of safely handling occasional encounters with singular configurations as a last resort to prevent loss of control. A thorough treatment of singularity handling is considered outside the scope of the present work. 

Instead, for the singular conditions $\|\boldsymbol{f}\| = 0$ and $\boldsymbol{v}_a~\parallel~\boldsymbol{f}$, we adopt heuristic strategies similar to those proposed in \cite{lu2024}, by setting the commanded $\boldsymbol{b}_y$ equal to its measured value at the previous time step. This is applied within a neighborhood of the singularity, where solutions become ill-conditioned. For near-hover conditions, defined for $\| \boldsymbol{v}_a \|<~1~\mathrm{m/s}$, we employ the uncoordinated differential flatness transform of \cite{tal2022} choosing the yaw angle reference as
$
\psi~=~\mathrm{atan2}\!\left(
\boldsymbol{i}_y^\top \boldsymbol{v}_a,\;
\boldsymbol{i}_x^\top \boldsymbol{v}_a
\right)
$ to satisfy the coordinated flight design criterion under the level flight assumption.

\subsection{Extension to dual-motor tailsitters}
To extend the formulation to dual-motor tailsitters, the different dynamic model in \cref{eq:fbxyz,eq:mbxyz} must be accounted for. This affects the computation of the angle $\theta_e$ and its derivatives, meaning the rotation from the intermediate frame $\mathcal{E}$ to the body-fixed frame $\mathcal{B}$. One such dynamic model is provided in \cite{tal2022}. The first rotation, from the inertial frame $\mathcal{I}$ to $\mathcal{E}$, remains unchanged. Finally, the recovery of the control input differs too, but this is also addressed in \cite{tal2022}.

\section{Controller design}
We employ the \gls{indi} controller structure presented in \cite{tal2022}. For completeness we briefly present it in the following. Let $\boldsymbol{p}_r(t) \in \mathbb{R}^{3}$ denote the reference trajectory, and $\boldsymbol{v}_r = \dot{\boldsymbol{p}}_r$, and $\boldsymbol{a}_r = \ddot{\boldsymbol{p}}_r$. The trajectory-tracking control law is
\begin{equation}
\label{eq:ac_traj}
    \boldsymbol{a}_c = \boldsymbol{a}_r -K_v(\boldsymbol{v} - \boldsymbol{v}_r) - K_p(\boldsymbol{p} - \boldsymbol{p}_r),
\end{equation}
where $K_p$,~$K_v~\in~\mathbb{R}^{3\times3}$ are diagonal gain matrices with positive elements. The incremental acceleration control law is 
\begin{equation}
    \label{eq:indi_accel}
    \boldsymbol{f}_c = (\boldsymbol{a}_c - \boldsymbol{a}_{f}) + \boldsymbol{f}_f,
\end{equation}
where $\boldsymbol{a}_f$ is the low-pass filtered acceleration feedback, and $\boldsymbol{f}_f$ is obtained from \cref{eq:fbxyz} using the filtered control input signal. Subsequently, $\boldsymbol{f}_c$ is used with \cref{eq:flatness:R} to compute the attitude control signal $R_{i,c}^b$. Let $\boldsymbol{q}_c$ be its equivalent quaternion representation. The quaternion error is obtained with $\boldsymbol{q}_e~=~\boldsymbol{q}_c \otimes \boldsymbol{q}^*$ \cite{fresk2013}.
The attitude control law is
\begin{equation}
    \label{eq:att_controller}
    \dot{\boldsymbol{\Omega}}_c = \dot{\boldsymbol{\Omega}}_\textrm{ff}     -
    K_\Omega(\boldsymbol{\Omega} - \boldsymbol{\Omega}_\textrm{ff}) -K_q\boldsymbol{q}_{e,v},
\end{equation}
where $\boldsymbol{q}_{e,v}$ is the vector part of the quaterion error, $K_q$,~$K_\Omega~\in~\mathbb{R}^{3\times3}$ are diagonal gain matrices with positive elements, and $\boldsymbol{\Omega}_\textrm{ff}$ and $\dot{\boldsymbol{\Omega}}_\textrm{ff}$ are the feedforward angular rate and angular acceleration signals computed by \cref{eq:flatness:Omega} and \cref{eq:flatness:Omegadot} respectively. The angular acceleration control incremental law is
\begin{equation}
    \boldsymbol{m}_c = (\dot{\boldsymbol{\Omega}}_c - \dot{\boldsymbol{\Omega}}_f) + \boldsymbol{m}_f,
\end{equation}
where $\dot{\boldsymbol{\Omega}}_f$ is the low-pass filtered angular acceleration feedback, and $\boldsymbol{m}_f$ is obtained from \cref{eq:mbxyz} using the filtered control input signal. Finally, $\boldsymbol{m}_c$ is used with \cref{eq:flatness:control_input} to calculate the commanded control input signal $\boldsymbol{u}_c$.

\section{Experimental study}
\label{sec:experimental_study}
For the experiments, we use the Parrot Swing drone in an indoor arena with a space of $10 \times 10 \times 5~\mathrm{m}$, equipped with a motion-capture system. The flight control algorithm is implemented in the Paparazzi autopilot framework and runs directly on the existing drone hardware. The objective is to demonstrate the correct operation of the proposed framework, particularly in scenarios where the uncoordinated differential flatness-based method becomes unsuitable.

We note that the platform is significantly underpowered and, together with the restricted flight space, makes trajectory generation challenging. Trajectories must be designed carefully to avoid actuator saturation while still achieving the desired objective. Trajectory optimization is considered outside the scope of the present work, thus trajectories are parameterized analytically and tuned manually.

\subsection{Half loop}
First, we consider a half loop maneuver in the East-Down plane consisting of three phases: a $4~\mathrm{m}$ straight line entry segment parameterized by a polynomial, a half-loop segment of radius $1.5~\mathrm{m}$ generated by integrating a smoothly increasing turning angle profile with prescribed speed, and a final straight line exit segment symmetric to the entry one. The complete trajectory is four times continuously differentiable. To avoid actuator saturation during the most demanding phase of the maneuver, namely the entry into the half loop, the entry speed is set to $2~\mathrm{m/s}$. The vehicle subsequently accelerates and exits the half loop at $3.2~\mathrm{m/s}$, before decelerating to a stop.

\begin{figure}[t]
    \centering
    \captionsetup[subfloat]{skip=0pt}
    \subfloat[Trajectory tracking with frame $\mathcal{B}$ axes $\boldsymbol{b}_x$ (blue), $\boldsymbol{b}_y$ (green), and $\boldsymbol{b}_z$ (red), shown at $0.5\,\mathrm{s}$ intervals.  \label{fig:immel_3D}]{
        \includegraphics[width=0.7\columnwidth]{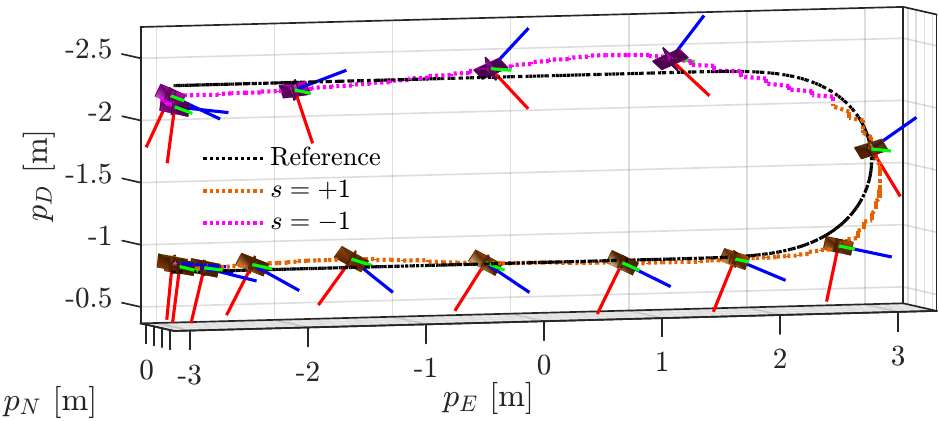}
    }
    \vspace{0.01cm}
    \captionsetup[subfloat]{skip=0pt}
    \subfloat[Commanded specific force; components and norm. \label{fig:immel_fcmd}]{
        \includegraphics[width=0.8\columnwidth]{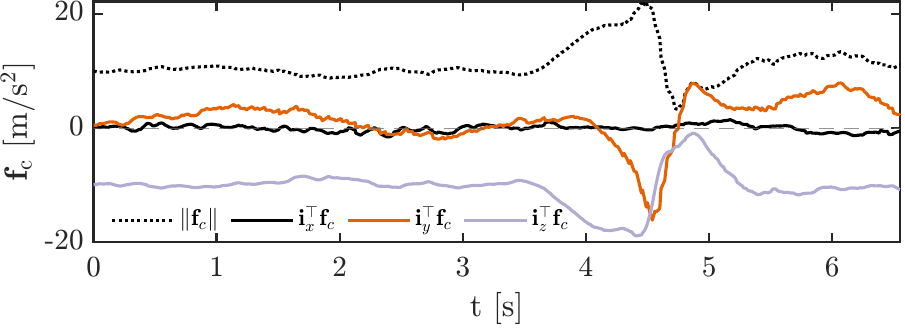}
    }
    \vspace{0.01cm}
    \captionsetup[subfloat]{skip=0pt}
    \subfloat[Attitude tracking. Euler angles (Z-X-Y) only for visualization.\label{fig:immel_att}]{
        \includegraphics[width=0.8\columnwidth]{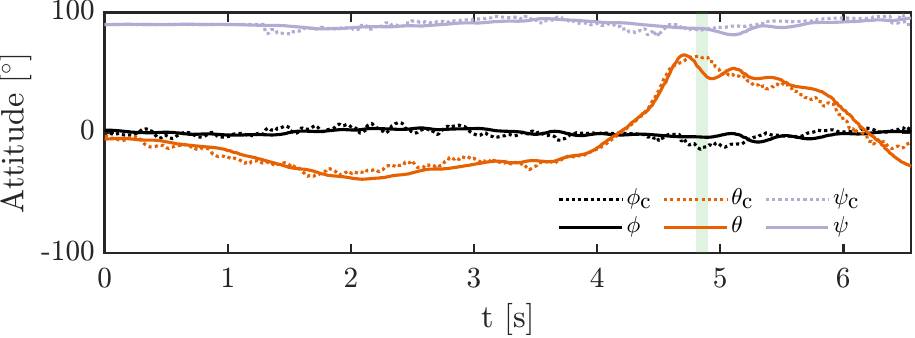}
    }
    \caption{Half-loop maneuver.}
\end{figure}

\Cref{fig:immel_3D} shows the reference and actual trajectories of the drone. The part of the actual trajectory highlighted in magenta is where the cross product $\boldsymbol{v}_a \times \boldsymbol{f}$ would normally undergo a sign change. However, the multiplication by $s$ in \cref{eq:by} ensures continuity of the commanded attitude. The tracking error norm remains below $0.4~\mathrm{m}$. \Cref{fig:immel_fcmd} shows the commanded specific force $\boldsymbol{f}_c$ generated by the \gls{indi} acceleration controller in \cref{eq:indi_accel}, which is then transformed to the commanded attitude $\boldsymbol{q}_c$ (\cref{fig:immel_att}) based on the proposed flatness transform in \cref{sec:diff_flatness}. Note that the use of Euler angles is only for visualization.

Achieving this maneuver with the uncoordinated flatness transform requires selecting the yaw angle as part of the flat output. Since the assumption of level flight no longer holds, the yaw angle cannot be defined in a meaningful way. In contrast, the proposed coordinated flatness transform is explicitly designed to provide a coordinated flight attitude command regardless of the flight path.

Note also that the component $\boldsymbol{i}_z^\top \boldsymbol{f}_c$ approaches zero at the exit of the half loop, highlighting another limitation of the uncoordinated flatness transform. As desctibed in \cite{tal2022} given the reference $\psi$ angle the algorithm subsequently computes the $\phi$ angle solving $\tan \phi~=~( -\sin\psi \boldsymbol{i}_x^\top \boldsymbol{f}_c~+~\cos\psi \boldsymbol{i}_y^\top \boldsymbol{f}_c  )/ \boldsymbol{i}_z^\top \boldsymbol{f}_c$. This expression becomes singular for $\boldsymbol{i}_z^\top \boldsymbol{f}_c = 0$, whereas the proposed formulation remains well defined. Such an operational scenario is shown in the green shaded region of \cref{fig:immel_att}.
 
\subsection{Half loop with cross-track motion}
Next, we consider a similar half loop maneuver of $3~\mathrm{m}$ radius and constant speed of $2.6~\mathrm{m/s}$ in the East-Down plane, with an additional cross-track motion along the North axis with a smooth speed profile reaching up to $2.5~\mathrm{m/s}$. The resulting reference speed reaches up to $3.6~\mathrm{m/s}$.

\begin{figure}[t]
    \centering
    \captionsetup[subfloat]{skip=0pt}
    \subfloat[Trajectory tracking with frame $\mathcal{B}$ axes $\boldsymbol{b}_x$ (blue), $\boldsymbol{b}_y$ (green), and $\boldsymbol{b}_z$ (red), shown at $0.5\,\mathrm{s}$ intervals.\label{fig:lateral_3D}]{%
        \includegraphics[width=0.67\columnwidth]{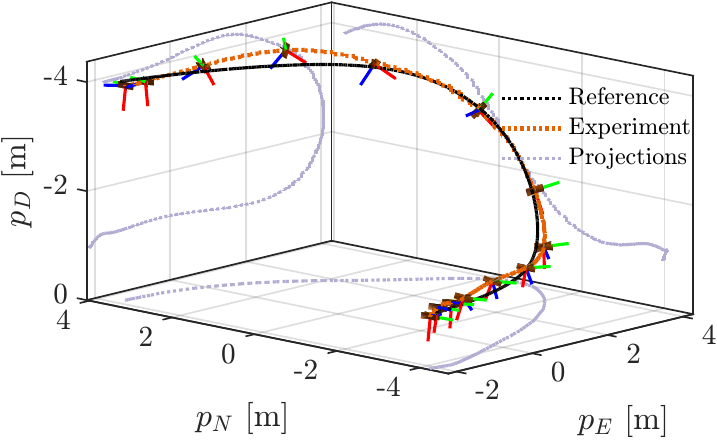}%
    }
    \vspace{0.01cm}
    \captionsetup[subfloat]{skip=0pt}
    \subfloat[Commanded specific force; components and norm.\label{fig:lateral_fcmd}]{%
        \includegraphics[width=0.8\columnwidth]{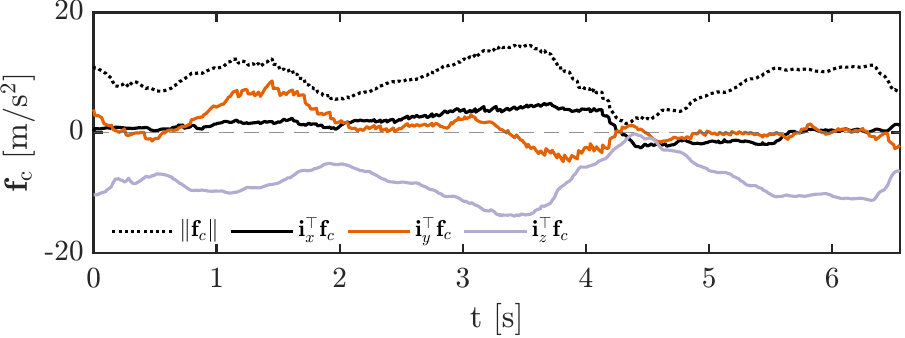}%
    }
    \vspace{0.01cm}
    \captionsetup[subfloat]{skip=0pt}
    \subfloat[Attitude tracking. Euler angles (Z-X-Y) only for visualization.\label{fig:lateral_att}]{%
        \includegraphics[width=0.8\columnwidth]{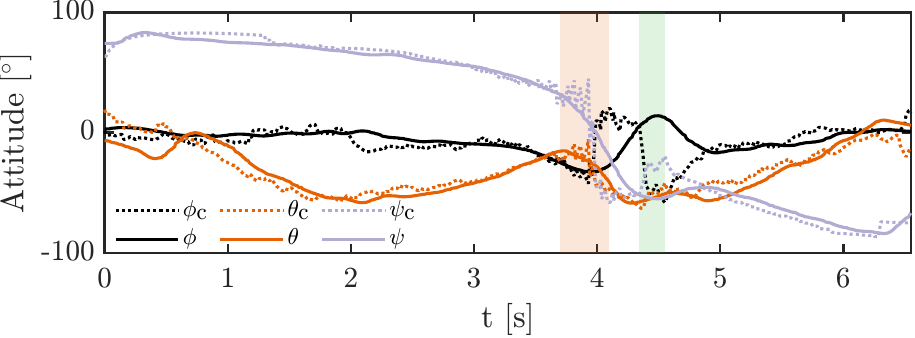}%
    }
    \caption{Half loop with cross-track motion maneuver.}
\end{figure}

The motivation behind this scenario is twofold. First, it represents a more generic three-dimensional maneuver. Second, it illustrates how a vehicle performing a half loop maneuver outdoors would simultaneously compensate for crosswind.
\Cref{fig:lateral_3D} shows the reference and actual trajectories of the drone. The tracking error norm remains below $0.4~\mathrm{m}$. \Cref{fig:lateral_fcmd} and \cref{fig:lateral_att} depict the commanded specific force $\boldsymbol{f}_c$ and the corresponding commanded attitude $\boldsymbol{q}_c$.

This experiment highlights the limitations of the uncoordinated flatness transform more clearly. In the previous case, one could argue that a zero yaw angle might be selected. Here, however, the presence of cross-track motion makes the a-priori definition of the yaw angle non-trivial for coordinated flight. Furthermore, the component $\boldsymbol{i}_z^\top \boldsymbol{f}_c$, becomes zero at $t=4.4~\mathrm{s}$ which, as discussed previously, introduces a singularity in the uncoordinated flatness transform. In contrast, the proposed transform remains well defined. This region is indicated by the shaded green area.

The shaded orange region corresponds to operation close to the singular condition $\boldsymbol{v}_a \parallel \boldsymbol{f}$, where the solution of the proposed transform becomes ill-defined. The transform exhibits some sensitivity but the spikes observed in the commanded attitude are primarily caused by the discrete and noisy velocity estimates of the positioning system. This condition further highlights the need for a trajectory optimization scheme, as noted previously. Nevertheless, the transform remains well-defined throughout the trajectory.

\section{Conclusion}

We have analytically derived the differential flatness transform for quadrotor tailsitters operating under the coordinated flight condition and discussed its extension to dual-motor tailsitters. The proposed formulation extends the applicability of a previous state-of-the-art control architecture by enabling coordinated flight, thereby allowing higher-speed maneuvers and addressing limitations of the earlier approach.

One difficulty encountered during the experiments was trajectory design. The trajectories were designed manually to satisfy the vehicle limitations while exciting specific flight regimes needed to demonstrate the desired behavior. Future work will focus on online trajectory generation that explicitly accounts for these constraints and on validating the complete framework in outdoor flight experiments. Finally, we also aim to explore combining the coordinated and uncoordinated differential flatness transforms within a unified framework, thereby enabling a broader range of aerobatic maneuvers.

\addtolength{\textheight}{-12cm}   % This command serves to balance the column lengths

%%%%%%%%%%%%%%%%%%%%%%%%%%%%%%%%%%%%%%%%%%%%%%%%%%%%%%%%%%%%%%%%%%%%%%%%%%%%%%%%

\bibliographystyle{IEEEtran}
\bibliography{references}

\end{document}